# Multi label classification of Artificial Intelligence related patents using Modified D2SBERT and Sentence Attention mechanism

Yongmin Yoo[1], Tak-Sung Heo[2], Dongjin Lim[2] and Deaho Seo[3]
[1] University of Auckland, Auckland, New Zealand
[2] AI R&D Team of AI R&D Group, NHN Diquest Seoul, Korea
[3] Graduate School of Information, Yonsei University, Seoul, Korea

Corresponding author: Deaho Seo (e-mail: seo_daeho@naver.com)

This work was supported in part by the Ministry of SMEs and Startups, South Korea under Grant S3319094

**ABSTRACT** Patent classification is an essential task in patent information management and patent knowledge mining. It is very important to classify patents related to artificial intelligence, which is the biggest topic these days. However, artificial intelligence-related patents are very difficult to classify because it is a mixture of complex technologies and legal terms. Moreover, due to the unsatisfactory performance of current algorithms, it is still mostly done manually, wasting a lot of time and money. Therefore, we present a method for classifying artificial intelligence-related patents published by the USPTO using natural language processing technique and deep learning methodology. We use deformed BERT and sentence attention overcome the limitations of BERT. Our experiment result is highest performance compared to other deep learning methods.

## I. INTRODUCTION

Patent automatic classification is a very important task in patent information management and patent knowledge mining [1]. If patents can be accurately classified, the flow of patents can reveal technological changes and trends. It is also very useful for technology roadmaps [2,3,4] technology forecasting [5,6,7], technology trend analysis [8,9,10], and furthermore, market marketing driven by technological change [11,12]. For this reason, automatic classification of patents has been research for a long time and conducted many ways. In particular, in a situation where technology changes rapidly and many new technologies are created these days, automatic classification of patents is a very important issue in terms of management.

Despite a lot of research, there have been many difficulties in classifying patents automatically. Due to the advent of deep learning and the development of natural language processing technique, it shows higher accuracy than past techniques in solving classification problems, but the results are not satisfactory. For this reason, patents are still mostly classified manually by patent attorneys and lawyers. In order to accurately classify patents, an understanding of legal terminology and technology is required, requiring a lot of effort and money from experts. And because it's a human job, it's error-prone and time-consuming. Therefore, research that solves the problem of automatically classifying patents has an even more important value. Therefore, we solve the problem of automatically classifying patents using natural language processing techniques.

The text classification problem we will experiment with is a representative downstream task of natural language processing, and research is being actively conducted in various fields [13]. In addition to classification of easily accessible texts such as news and books [14,15], more specialized and complex legal documents [16,17,18,19] and studies on the classification of patient medical records [20,21,22], etc. A lot of research in multi label classification is being done in various fields.

As the classification problem has emerged as an important problem in natural language processing, the multi-label classification problem in which various labels are applied to one description is also recognized as an important problem in natural language processing. For example, text with many complex elements, such as prescriptions, cannot be described with a single label. Therefore, one text data often has multiple labels. Multi-label classification is the task of attempting to classify these labels automatically.

A patent is a legal document that describes technology. A single patent often has multiple labels, as a single technology is often an aggregation of technologies with several sub-technologies combined. Therefore, patent classification is a multi-label classification problem. We try to solve the multi-label classification problem of predicting IPC codes using AI-related patents issued by the United States Patent and Trademark Office (USPTO) in 2019 and 2020. The International Patent Classification (IPC) is an internationally unified patent classification system. An IPC code indicating the technical field of the invention is affixed to every patent application. In order to assign an IPC code to each patent, it requires a lot of effort and cost to do this manually, as it requires an understanding of precise legal terminology and technology. Therefore, we propose a method to solve the multi-label classification problem, which is the task of automatically classifying multiple codes for one patent. This research describes a method for automatically assigning IPC codes using AI-related patent data in 2019 and 2020 announced by the USPTO as a dataset.

Our model used the previously proposed D2SBERT and Attention model [20]. The D2SBERT model is a model that solves the sequence max length problem of BERT. However, to increase the novelty of the research, we did not use the model as it is, but modified it to suit the characteristics of the patent.

The contributions of this research are as follows:

- BERT has a maximum sequence length limitation when using pretrained models, making it difficult to apply directly to relatively long documents. To overcome these disadvantages, we use the D2SBERT method, which is a BERT that can be applied to documents by embedding each sentence, and the Sentence Attention mechanism that captures important sentences appearing in documents.
- Artificial intelligence-related patents, which are very difficult documents composed of legal terminology and scientific technology, were automatically classified. Artificial Intelligence related technology has many patents created by the sum of various sub-technologies, making automatic labeling very difficult, but our method showed high accuracy than any other deep learning method.

This research is composed as follows. Section 2 describes related works. Section 3 describes the dataset. Section 4 explains the proposed methodology and details of the main components. Section 5 describes our research methodology. Finally, Section 6 describes the conclusion of this research and discusses future studies.

## II. LITERATURE RESEARCH

Prior research was divided into three categories. First, we investigated previous studies on patent classification, which is our main research. Second, we investigated previous studies on multi-label classification, one of the tasks of natural language processing. Thirdly, I wrote down the description of the IPC code corresponding to the label. Finally, the structure of the patent, which is the dataset used in the experiment, is explained.

### A. Patent Classification

As many technologies develop rapidly and the importance of intellectual property rights increases, patents become more and more important. For this reason, various studies have been conducted in the past to automatically classify patents. Li et al. proposed a DeepPatent model that classifies IPC codes using word embedding and Convolutional Neural Network. Embedding the patent title and abstract was used for CNN learning, and an f1 score of 43% was recorded in the problem of classifying 606 ICP codes [22]. Lee et al. presented a patentBERT model that classifies IPC codes by fine-tuning the pretrained BERT with a patent dataset. BERT was trained using IPC, Title, and Abstract among patent datasets, and showed an f1 score of 65.87% [23]. H Bekamiri et al. proposed a method of classifying CPC codes using SBERT (Sentence BERT) and KNN to solve the problem of slow prediction speed of the existing BERT model. We trained SBERT using CPC and Claim and achieved an F1 score of 66.48% for 663 labels with classification using KNN [24]. R Henriques et al. consists of 30,000 patent data in Portuguese and conducted research to classify 124 IPC categories with various ML algorithms. LinearSVC et al. used as a baseline and CNN, BiLSTM, and BERT models were compared. As a result of the study, the transfer learning model of the BERTimbau model recorded 63.6%, about 2.8% higher than the baseline [25]. A Haghighian Roudsari et al. conducted a study to distinguish between 544 USPTO-2M IPC Codes and 96 M-patent IPC Codes. Pre-trained BERT, XLNet, RoBERTa, and ELECTRA were used as deep learning models, and XLNet showed the best performance with an F1-score of 63.33% in the USPTO-2M dataset [26].

### B. Multi-label classification

With an era in which there are many documents with complex characteristics and the number of documents itself is rapidly increasing. For these reasons, automatic multi-label classification is an important task in the field of natural language processing. Before deep learning era, many research are based on statistical machine learning such as Term Frequency-Inverse Document Frequency (TF-IDF), naïve Bayes Classifier and Support Vector Machines (SVMs) etc [27,28]. Wenqian et al [29] proposed f-k-NN (fuzzy k-NN) algorithm for the improvement of decision rule and design to improve classification performance in uneven class distribution datasets. In Dino et al [30], used naive Bayes method at raw text data vectorization and then used SVM for classify the documents to the right category.

As deep learning research continues to show good performance various fields, in NLP, also many experiments are being conducted and get outgoing result [20,31,32,33]. Baumel et al [31] proposed Hierarchical Attention bidirectional Gated Recurrent Unit (HA-GRU), a hierarchical approach to tag a document by identifying the sentences relevant for each label. Heo et al propose a model based on bidirectional encoder representations from transformer (BERT) using the sequence attention method [20]. Adhikari et al propose a fine-tuned BERT model for multi-label classification [32]. Liu et al. propose a Label-Embedding Bi-directional Attentive model to improve the performance of BERT's text classification framework for solving multi label classification problem [33].

### C. International Patent Classification (IPC)

| IPC hierarchy | Description |
|---|---|
| B | : Section B - Performing operations; transporting |
| B82 | : Nanotechnology |
| B82Y | : Specific uses or applications of nano-structures; MEASUREMENT OR ANALYSIS OF NANOSTRUCTURES; MANUFACTURE OR TREATMENT OF NANOSTRUCTURES; |
| B82Y 20/00 | : Nanooptics, e.g. quantum optics or photonic crystals |

Figure 1. Example of IPC hierarchy

A patent is a legal document and right to protect an inventor's technology. All issued patents are protected by each national government. Patent is a legal document with technological information such as patent number, title, abstract, inventor, claim, International Patent Classification (IPC) code, Cooperative Patent Classification (CPC) and so on. Therefore, we must think patent with one technology, not only just one legal document. The IPC is a hierarchical patent classification established by the Strasbourg Agreement 1971. It has been administered by World Intellectual Property Organization (WIPO). IPC code has a 5-level hierarchical structure of section, class, subclass, main group, and subgroup. A section represents the top level of the IPC hierarchy and is divided into 8 sections. First level of IPC code consists of A to H. They are Human Necessities (A), Performing Operations, Transporting (B), Chemistry, Metallurgy (C), Textiles, Paper (D), Fixed Constructions (E), Mechanical Engineering, Lighting, Heating, Weapons (F), Physics (G), and Electricity (H). The subclass is followed by main group of IPC code. The second level, called classes, is the subdivision of a section, followed by a section symbol followed by two numbers. The third level is called a subclass and is followed by a class symbol followed by a capital letter. The fourth level is called the main group and consists of 1 to 3 numbers followed by a forward slash (/) and the number 00. Lastly, the subgroup forms a detailed development item under the main group and means that at least two numbers other than 00 are added after the main group and a slash.

Fig 1 shows example of hierarchical structure of IPC codes The IPC code represents a technical classification term that represents an invented topic [34]. The IPC code represents a technical classification term that represents an invented topic. In our research, we use 3-depth of IPC code such as B82Y in Fig. 1. Most IPC code hierarchy of patent documents is based on this code level.

### III. DATASET

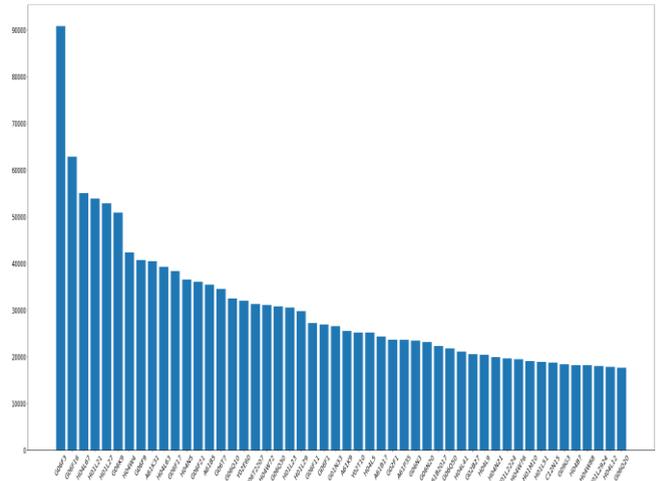

Figure 2. Statistics of 50 most frequent IPC codes

The U.S. Patent and Trademark Office, an agency of the U.S. Department of Commerce that issues patents to inventors and companies for trademark registration and inventions for identifying products and intellectual property rights, called the USPTO, published artificial intelligence-related patents from 2016 to 2020 on the USPTO site [35]. The domains of published artificial intelligence-related patents consist of machine learning, NLP, CV, speech, knowledge processing, AI hardware, evolutionary computation, planning, and control. We crawled patents from Google patent based on the patent numbers of the AI-related patent dataset provided by the USPTO. We use 5tables 'publication number', 'title', 'abstract', and 'IPC code', ''description' our experimental dataset. Reflecting the characteristics of the rapidly changing ai market, we use data from the last two years among the publicly available artificial intelligence datasets as a dataset. In addition, we use only sections, classes, and subclasses as IPC codes to secure the disadvantage that IPC codes are too subdivided and cannot be unified. Also, since there are many different IPC codes, we solve the multi-label classification problem using only the top 50 IPC codes. The number of data used in this study is 632,105, and the experiment is conducted by dividing it into training set: validation set: test set = 8: 1: 1.

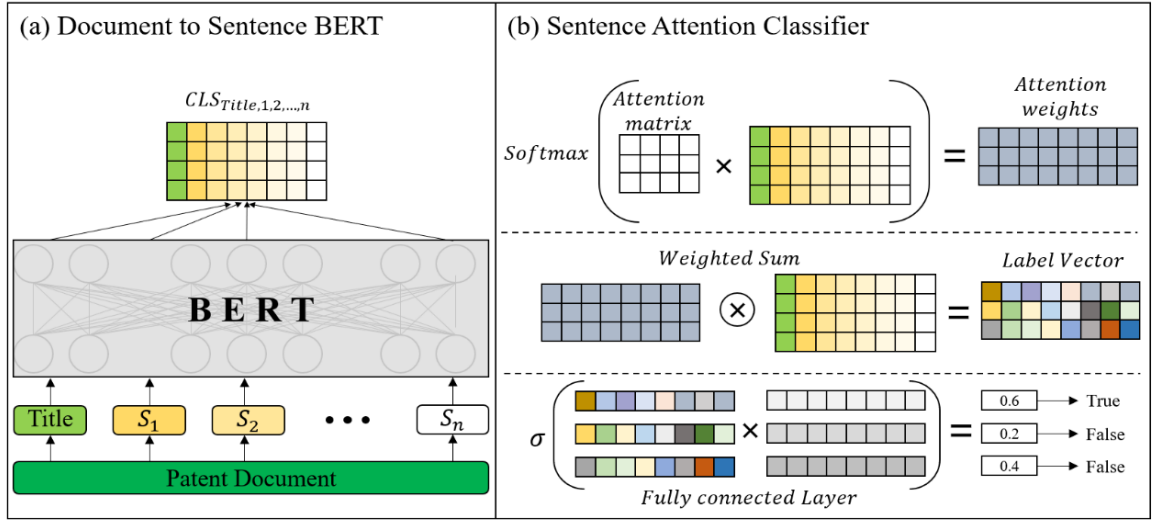

Figure 3. Architecture of proposed model

## IV. METHODOLOGY

In our research, we modify and use the D2SBERT model proposed earlier. The D2SBERT model proposed a multi-labeling classification model for medical codes using the text part of the MIMIC-III data, an American critical care data [20]. However, there are many risks in using Sequence Attention proposed in D2SBERT. When cutting text into specific sequence units, the middle part of a text may be cut off or omitted, so the positional embedding value considering the temporal order of sentences cannot be properly reflected. For this reason, our model uses sentence attention, a method that considers the temporal order of sentences. In addition, we propose a model of Sentence Attention that recognizes the relational contents of documents. As shown in Figure 3, the model is divided into D2SBERT, which extracts and processes all information in a sentence as much as possible, and SAC (Sentence Attention Classifier), which predicts labels by reflecting only important information.

### A. Document to Sentence BERT(D2SBERT)

To proceed with the multi-label text classification task, BERT adds two special tokens at the beginning of the sentence [CLS] and the end of the sentence [SEP] to the dataset text and uses them as input values. Existing BERT models suffer from a specified maximum sequence length problem. Typically, inputs are truncated when fine-tuning BERT with long sequences. However, it is not suitable for patent code predicates with missing data, especially requiring a detailed understanding of the sentence. To solve this problem, D2SBERT first broke the entire document into sentences using the Spacy library's Sentence tokenizer in Python. Each segmented sentence is used as an input to the BERT model. To extract information about each sentence, we use [CLS] tokens extracted via BERT. We define the [CLS] token extracted through each input sentence as $CLS_i$. When $CLS_i$ has the dimension of $\mathbb{R}^h$ and the number of sentences is $k$, the document representation vector $(D)$ containing all sentence information is $\mathbb{R}^{h \times k}$. $D$ is represented by Equation (1).

$$D = [CLS_1; CLS_2; ...; CLS_k;] \qquad (1)$$

### B. Sentence Attention Classifier

Sentence attention decides important parts in the document by utilizing the relationship of the sentences. In the sentence attention classifier, the model extracts the important information from $D$ by using the sentence attention mechanism. When the total number of labels to be classified is c, $D$ and sentence attention matrix (S) having the dimension of $\mathbb{R}^{h \times c}$ are utilized to generate attention weights (α). α is obtained as in equation (2).

$$\alpha = softmax(tahn(SD^T)) \qquad (2)$$

In Equation 2, α is calculated with the *softmax* function. The label representation vectors ($L = [l_1, l_2, l_3, ...., l_n]$) are calculated using attention weight, which extracts the important information from a sentence and document. $l_i$ has the dimension of $\mathbb{R}^{h \times 1}$. Equation (3) shows the process of calculating $l_i$.

$$l_i = \sum_{j=1}^{n} a_{ij} D_j \qquad (3)$$

The information needed to determine whether the target label result is true or false is contained in $l_i$. After each $l_i$ is connected to another different fully connected layer MLP, the final score of labels is calculated using sigmoid function (σ). The final score about each target label is calculated as equation (4, 5).

$$MLP_i(x) = \sum_{j=1}^{h} W_j x_j + b \quad (4)$$

$$Score_i = \sigma(MLP_i(l_i)) \quad (5)$$

$Score_i$ has between 0 and 1. Finally, true or false of the label is determined by equation (6).

$$Predict_i^{IPC} = \text{if True: } Score_i > 0.5 \quad (6)$$
$$= \text{Else False: otherwise}$$

## V. EXPERIMENT

The performance of the model proposed in this research is compare with CNN, LSTM, Bi-LSTM, BERT-head, BERT-tail and D2SBERT models. The performance evaluation metrics of our model is evaluated with F1 macro and F1 micro values.

### A. Evaluation Metrics

We use two standard metrics to measure and compare the performance of our models. We use macro-average F1 and micro-average F1. Two standard metrics are calculated using precision and recall. And precision and recall are calculated using true-positive (TP), false-positive (FP), and false-negative (FN).

Macro average F1 calculates metrics for each class independently. It's a standard metric that averages those values after that. The macro average F1 is calculated according to the equation (7, 8).

$$Precision_{macro} = \frac{1}{C}\sum_{c=1}^{c} \frac{TP_c}{TP_c + FP_c} \quad (7)$$

$$Recall_{macro} = \frac{1}{C}\sum_{c=1}^{c} \frac{TP_c}{TP_c + FN_c} \quad (8)$$

The fine-average F1 aggregates the contributions of all classes. After that, the average is taken to calculate the metric. The micro-averaged F1 is calculated by the equation (9, 10).

$$Precision_{micro} = \frac{\sum_{C=1}^{C} * TP_c}{\sum_{C=1}^{C} * (TP_c + FP_c)} \quad (9)$$

$$Recall_{micro} = \frac{\sum_{C=1}^{C} * TP_c}{\sum_{C=1}^{C} * (TP_c + FN_c)} \quad (10)$$

Lastly, macro-average F1 and micro-average F1 are calculated as harmonic averages of precision and recall, as shown in equation (11).

$$F1_{score} = 2 * \frac{Precision * Recall}{Precision + Recall} \quad (11)$$

### B. Result

| Model | F1 | |
|---|---|---|
| | *Macro* | *Micro* |
| CNN | 0.6472 | 0.6520 |
| LSTM | 0.6733 | 0.6682 |
| Bi-LSTM | 0.6803 | 0.6891 |
| BERT-head | 0.7043 | 0.7083 |
| BERT-tail | 0.6925 | 0.6972 |
| D2SBERT | 0.7094 | 0.7114 |
| **Proposed Model** | 0.7234 | 0.7341 |

Table 1. Result of Experiment

Our research compared CNN and LSTM, Bi-LSTM and BERT-head, BERT-tail model, D2SBERT and our proposed model. The comparison results are recorded in Table 1. above. In the results, our model performed much better than the classic deep learning models CNN, LSTM, and Bi-LSTM. And it showed better performance than the head and tail of the BERT model, which shows the best performance in natural language processing. In addition, the previously proposed D2SBERT model has lower performance than our proposed model. In other words, we have overcome the problem that the middle part of our model text can be truncated or omitted and the positional embedding value considering the temporal order of sentences is not properly reflected. In addition, we proved that Sentence Attention, which considers the temporal order of sentences proposed by our model, is a very effective method.

## VI. CONCLUSION AND FUTURE WORKS

The final part of this research was divided into conclusion of this research and future research. The conclusion section describes the final conclusions and advantages of our research, and the future research section suggests the direction for future research through our research.

### A. Conclusion of this research

Due to the nature of patents made up of legal term and technological term, it is very difficult for artificial intelligence to automatically classify patents. Even if prior studies are confirmed, they do not show high accuracy. Therefore, lawyers and patent attorneys still classification manually. Patent classification is time consuming and expensive. Therefore, automatically classifying patents is an incredibly challenging task. Therefore, we conducted research to automatically classify patents using IPC codes as labels. The IPC code is an important code that represents the technology possessed by each patent. It is used to arrange and disseminate patent documents in an orderly manner for easy access to right information and is used to evaluate technological development in various fields and create statistics. Also, it is used prior technology research. Therefore, it is very important to classify IPC codes quickly and accurately. In particular, in the field of technology management, it is a very important factor in one field of technology management to accurately classify patents, such as attempting technology prediction to predict future technology by analysing the flow of patents and making investment proposals based on patent analysis. Also, from the point of view of the natural language processing field, multi-label classification that automatically labels one text containing various information is very valuable. Our research solved problems in the field of technology management and natural language processing at the same time. An experiment was conducted using an artificial intelligence-related patent dataset officially announced by the USPTO, and the proposed method obtained better performance than other deep learning methods. In particular, it is very noble to automatically classify IPC codes of AI-related patents, a new technology that has recently increased in value. Due to the nature of artificial intelligence patents, one technology is composed of the sum of several new sub-technology, which makes the classification problem more difficult and complicated. This complex and difficult AI patent classification problem was solved using a modified model of the previously proposed D2SBERT model and a text attention model. The D2SBERT model has a disadvantage that when cutting text into specific sequence units using Sequence Attention, the middle part of the sentence is cut off or omitted, so that the positional embedding value considering the temporal order of the sentence may not be properly reflected. For this reason, our model performed better using the modified method of D2SBERT and Sentence Attention, which considers the temporal order of sentences.

### B. Future works

Future research can be confirmed in two aspects: technology management and natural language processing.

Patent automatic classification is a very important task in patent information management and patent knowledge mining. Many problems in the field of technology management can be solved by automatically and accurately classifying patents. If patents can be accurately classified, changes and trends in technology can be identified through the flow of patents in terms of technology management. In particular, our research is very valuable in the real world because the dataset of this experiment is a patent dataset related to AI, which is rapidly growing recently. It is also very useful for artificial intelligence technology roadmap, artificial intelligence technology prediction, artificial intelligence technology trend analysis, and furthermore, artificial intelligence market marketing according to technological changes. Again, it is very valuable to have a good understanding of one of the biggest and most sophisticated technologies of our time: artificial intelligence.

We solved one of the big problems of natural language processing, the multi-label classification problem. Our research overcomes the limitations of the previously proposed D2SBERT model. Our proposed model considers the problem that the middle part of the text may be cut off or omitted and the positional embedding value considering the temporal order of the sentence is not properly reflected. In addition, we proved that Sentence Attention, a method that considers the temporal order of sentences suggested by our model, is a very valid method. From a natural language processing point of view, our model can be used in multiple domains. Subsequent research may attempt to apply it to areas of complex and lengthy textual data, such as classifying case law or classifying medical data.

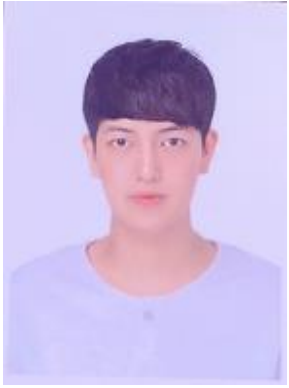

**Yongmin Yoo** was born in 1991 in Busan, South Korea. He received the master's degree in industrial engineering from Inha university, Korea. He is currently PhD student in University of Auckland, New Zealand.

He has experience working as a natural language processing researcher at NHN which is one of the biggest companies in Korea.

His research interests are in datamining especially textmining, deep learning, machine learning and natural language processing.

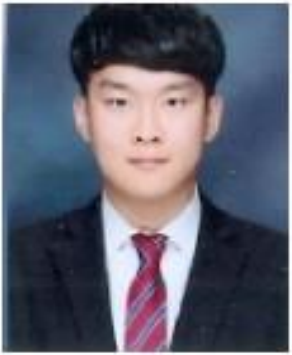

**Tak-sung Heo** was born in 1994 in Gyeonggi do, he also and receive his B.S degree in biology and M.S degree in engineering from Hallym University. He is currently working on natural language processing at NHN Diquest. His research interests are in deep learning, data mining, and natural language processing.

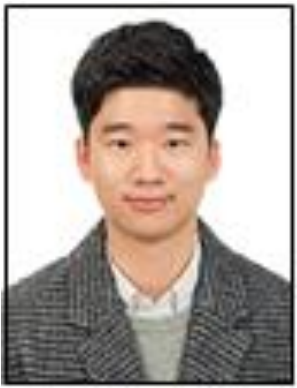

**Dongjin Lim** was born in 1992 in Seoul, he also received the master's degree in big data science. He is currently work as AI researcher in NHN, Korea. He is interested in big data analysis using deep learning and machine learning, especially vision tasks and natural language processing tasks.

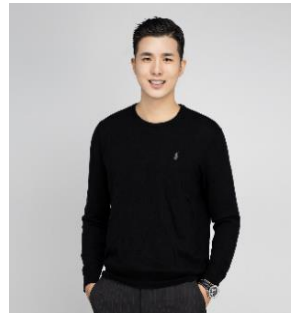

**Daeho Seo** was born in 1991 in Seoul, South Korea. He received his B.S degree in information systems and his M.S degree in industrial engineering from Hanyang University. He is Ph.D candidate from Yonsei University Graduate School of Information.

He has experience working as a big data researcher at the Korea Advanced Institute of Science and Technology and the Korea Electronics Technology Institute. Currently He is serving as CEO of Elesther. His research interests include text mining, vision-based anomaly detection, start factory, and e-commerce solution.

He has the experience of receiving the Best Young Entrepreneur Award from the Ministry of SMEs and Startups in Korea. He also has experience in publishing 8 books related to artificial intelligence and big data.